\documentclass[runningheads]{llncs}
\usepackage{cite}
\usepackage{amsmath,amssymb,amsfonts}
\usepackage{mathtools}
\usepackage{epsfig}
\usepackage[mathscr]{euscript}
\usepackage{color}
\usepackage{booktabs} % For formal tables
\usepackage{stmaryrd} % Kruskal operator
\usepackage{bbm} % for digital payoff
\usepackage{blindtext, comment}
\usepackage{rotating} % to rotate some text
\usepackage{tikz}
\usepackage{standalone}
\usetikzlibrary{arrows, patterns}
\usetikzlibrary{decorations.pathreplacing,angles,quotes}
\usepackage{framed, standalone}
\usepackage{graphicx}
\usepackage{capt-of}
\usepackage{balance}
\usepackage{makecell} % break line in a table cell
\usepackage{afterpage}
\usepackage[linesnumbered,lined,boxed,commentsnumbered,ruled,vlined]{algorithm2e}
\usepackage{setspace}
\usepackage{textcomp}
\usepackage{xcolor}
\def\BibTeX{{\rm B\kern-.05em{\sc i\kern-.025em b}\kern-.08em
    T\kern-.1667em\lower.7ex\hbox{E}\kern-.125emX}}
\begin{document}
\title{SynGAN: Towards Generating Synthetic Network Attacks using GANs
%Q-Learning for Retail Banking
}
\titlerunning{SynGAN: Towards Generating Synthetic Network Attacks using GANs}
%
%\author{\Large \textit{Anonymous} \vspace{1.85cm}}
\author{Jeremy Charlier \inst{1}\thanks{Part of this work was done at the Internet Real-Time Lab at Columbia University.}
\and
Aman Singh  \inst{2}
\and
Gaston Ormazabal \inst{3}
\and
Radu State \inst{1}
\and
Henning Schulzrinne\inst{3}}

%\authorrunning{Anonymous}
\authorrunning{J. Charlier et al.}
\institute{University of Luxembourg, L-1855 Luxembourg, Luxembourg  \\
\email{\{name.surname@\}@uni.lu}
\and 
Palindrome Technologies, New Jersey, USA\\
\email{aman.singh@palindrometech.com}
\and
Columbia University, New York NY 10027, USA  \\
\email{\{gso,hgs\}@cs.columbia.edu}
}
\maketitle

\begin{abstract}
The rapid digital transformation without security considerations has resulted in the rise of global-scale cyberattacks. The first line of defense against these attacks are Network Intrusion Detection Systems (NIDS). Once deployed, however, these systems work as blackboxes with a high rate of false positives with no measurable effectiveness. There is a need to continuously test and improve these systems by emulating real-world network attack mutations. We present SynGAN, a framework that generates adversarial network attacks using the Generative Adversial Networks (GAN). SynGAN generates malicious packet flow mutations using real attack traffic, which can improve NIDS attack detection rates. As a first step, we compare two public datasets, NSL-KDD and CICIDS2017, for generating synthetic Distributed Denial of Service (DDoS) network attacks. We evaluate the attack quality (real vs. synthetic) using a gradient boosting classifier.
\end{abstract}
\section{Introduction} \label{sec::intro}
Advancements in Machine Learning (ML) methods have given malicious actors new cyber-offense tools, resulting in high-volume and complex attacks. The first line of defense against these attacks are Network Intrusion Detection System (NIDS) that can learn network activity patterns by monitoring traffic, and raise an alarm when malicious traffic is encountered. These systems typically rely on both static attack signatures and dynamic behavior learning methods using data and time variations. These detection methods can be potentially evaded by attack mutations and complex learning algorithms \cite{bao2017shall}. They can also have high rates of false positives with no measurable effectiveness \cite{ficke2018characterizing, sommeroutside}. There is a need to continuously test, improve, and evolve NIDS models using real-world network attack mutations that can enhance both accuracy and complexity of attack pattern detection.  %Additionally, a recent trend among ML specialists is to use Deep Neural Networks (DNNs) to solve complex tasks or to improve the results accuracy on specific applications. %We recall a neural network is an elementary unit receiving a weighted input \cite{goodfellow2016deep}. A non linear function, or activation function, such as the sigmoid function or the Rectified Linear Unit \cite{nair2010rectified}, is used to transform the input signal to an output signal. A deep learning neural network is a large collection of neural networks glued together through different layers  \cite{goodfellow2016deep}. 
Deep Learning Neural Networks \cite{goodfellow2016deep} have been previously applied to improve NIDS \cite{yin2017deep} systems. The most recent Gradient Penalty-Wasserstein Generative Adversial Networks (GP-WGAN) \cite{gulrajani2017improved} can address both complexity and high-quality of synthetic network traffic flow generation \cite{ring2019flow}. \\

\begin{figure}[b!]
  \begin{center}
  \begin{tikzpicture}[scale=0.6, transform shape, line cap=round,line join=round,>=triangle 45,x=1cm,y=1cm]

% X rectangle
\draw [line width=0.25pt] (3,7) rectangle (6,9);
\fill [gray, opacity=.4] (3,7) rectangle (6,9);
\draw (4.15,8.3) node[anchor=north west] {\Large{$\mathcal{X}$}};

% Z rectangle 
\draw [line width=0.25pt] (-2.5451,7.55) rectangle (1,8.5);
\fill [gray, opacity=.4] (-2.5451,7.55) rectangle (1,8.5);

% tilde X rectangle
\draw [line width=0.25pt] (-7.5,7) rectangle (-4.5,9);
\fill [gray, opacity=.4] (-7.5,7) rectangle (-4.5,9);

% lines between X and Z rectangles
\draw [line width=0.25pt] (-4.5,9)-- (-2.55,8.55);
\draw [line width=0.25pt] (-4.5,7)-- (-2.55,7.55);

% lines between Z and tilde X rectangles
\draw [line width=0.25pt] (1,8.5)-- (3,9);
\draw [line width=0.25pt] (1,7.5)-- (3,7);

% X rectangle
\draw [line width=0.25pt] (-7.5,3) rectangle (-4.5,5);
\fill [gray, opacity=.4] (-7.5,3) rectangle (-4.5,5);

% Z rectangle 
\draw [line width=0.25pt] (-1.9951,3.5) rectangle (0.5049,4.5);
\fill [gray, opacity=.4] (-1.9951,3.5) rectangle (0.5049,4.5);
\draw (-6.4,4.3) node[anchor=north west] {\Large{$\mathcal{Z}$}}; 

% tilde X rectangle
\draw [line width=0.25pt] (3,3) rectangle (6,5);
\fill [gray, opacity=.4] (3,3) rectangle (6,5);

% lines between X and Z rectangles
\draw [line width=0.25pt] (-4.5,5)-- (-2,4.5);
\draw [line width=0.25pt] (-4.5,3)-- (-2,3.5);

% lines between Z and tilde X rectangles
\draw [line width=0.25pt] (0.5,4.5)-- (3,5);
\draw [line width=0.25pt] (0.5,3.5)-- (3,3);

% network traffic rectangle
\draw [line width=0.25pt] (8,7) rectangle (11,9);
\fill [gray, opacity=.2] (8,7) rectangle (11,9);

% gradient boosting ids rectangle
\draw [line width=0.25pt] (8,3) rectangle (11,5);
\fill [gray, opacity=.2] (8,3) rectangle (11,5);

% labels
\draw (-8,9.55) node[anchor=north west] {{PREDICTED LABELS}};
\draw (-7.2,8.3) node[anchor=north west] {Fake / True };
\draw (-2.,8.25) node[anchor=north west] {\scriptsize{DISCRIMINATOR}};
\draw (2.6,9.55) node[anchor=north west] {{ORIGINAL ATTACKS}};

\draw (-7.55,2.95) node[anchor=north west] {{RANDOM NOISE}};
\draw (-1.85,4.2) node[anchor=north west] {\scriptsize{GENERATOR}};
\draw (2.4,3) node[anchor=north west] {{GENERATED ATTACKS}};

\draw (9.5,8) node[anchor=south] {NETWORK};
\draw (9.5,7.5) node[anchor=south] {TRAFFIC};

\draw (9.5,4.2) node[anchor=south] {GRADIENT};
\draw (9.5,3.7) node[anchor=south] {BOOSTING};
\draw (9.5,3.2) node[anchor=south] {EVALUATOR};

\draw (-7.55,5) node[anchor=north west] {$P_Z$};
\draw (-1.3,9.55) node[anchor=north west] {$D$}; 
\draw (-1.5,5.5) node[anchor=north west] {$G$}; 
\draw (3,5) node[anchor=north west] {$P_\mathcal{\widetilde{X}}=P_G$};
\draw (4.15,4.35) node[anchor=north west] {\Large{$\mathcal{\widetilde{X}}$}};

\draw [line width=0.25pt, dashdotted, -triangle 60] (4.5,5.3) -- (-0.5,7.2);
\draw [line width=0.25pt, -triangle 60] (-3.5,5) -- (1.5,5);
\draw [line width=0.25pt, -triangle 60] (1.5,9) -- (-3.5,9);

\draw [line width=0.5pt, dotted, -triangle 60] (6.5,8) -- (7.5,8);
\draw [line width=0.5pt, dotted, -triangle 60] (6.25,4) -- (7.75,6.5);
\draw [line width=0.5pt, dotted, -triangle 60] (9.5,6.5) -- (9.5,5.5);

\end{tikzpicture}
    \caption{SynGAN framework components and operational flows.} \label{fig::methodology}
  \end{center}
\end{figure}
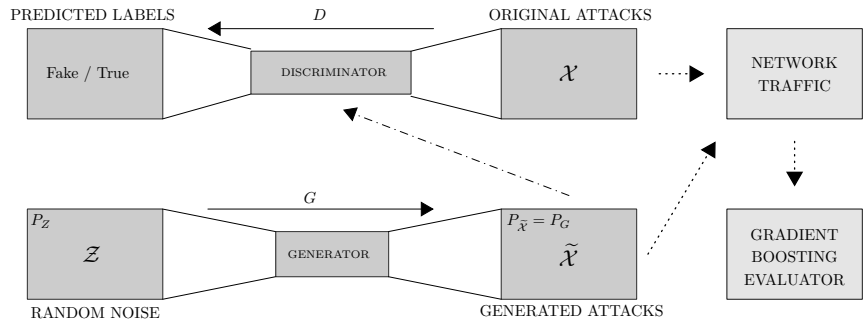

In this work, in contrast to \cite{ring2019flow}, we apply the GP-WGAN algorithm to specifically generate synthetic network attacks, using publicly available datasets. We present the SynGAN framework consisting of three components: the Generator, the Discriminator and the Evaluator, as shown in figure \ref{fig::methodology}. %The generator produces synthetic network attacks based on an initial random distribution. The discriminator provides feedback to the generator to improve the correctness of the adversarial attacks based on the observation of real network attacks. Additionally, 
The generated packets are measured against a quality benchmark, %The evaluator uses a gradient boosting binary classifier \cite{friedman2002stochastic} to measure synthetic packet quality. 
defined by the similarity between generated ``synthetic'' packets and real packets. We compare %and evaluate 
two real-world network datasets NSL-KDD \cite{tavallaee2009detailed} and CICIDS2017 \cite{sharafaldin2018toward} for synthetic network attack generation. Although NSL-KDD is a well-known network security dataset, it has short-comings in evaluating NIDS \cite{mchugh2000testing}. For our initial framework development and evaluation, however, the NSL-KDD dataset provides a good starting reference point. We evaluate the framework against the Distributed Denial of Service (DDoS) family of attacks that can have complex time variations.\\

%The paper is structured as follows. 
In section \ref{sec::background}, we describe the methodology including theoretical background of GP-WGANs, the DDoS taxonomy, and the SynGAN framework. In section \ref{sec::experiments}, we highlight experimental results. We conclude in section \ref{sec::conclusion} by addressing promising directions for future work. 

% !!! NEW SECTION !!! %
% !!! =========== !!! %
\section{Methodology} \label{sec::background}
In this section, we describe a brief theoretical formulation of GP-WGAN %based on the Kantorovich-Rubinstein duality summarized in \cite{villani2003topics}. 
followed by well known attacks from the DDoS taxonomy. Subsequently, we present the SynGAN framework and the generation of synthetic DDoS network attacks.

\subsection{Gradient Penalty Wasserstein GAN (GP-WGAN)}
The % in unsupervised learning, % has evolved significantly thanks to adversarial networks publications. %Variational Auto-Encoders (VAE) presented by Kingma et al. in \cite{kingma2013auto} have emerged as a well-established approach for synthetic data generation. We recall an AE is a neural network trained to copy its input manifold to its output manifold through a hidden layer. The encoder function sends the input space to the hidden space and the decoder function brings back the hidden space to the input space. Simultaneously, 
GAN \cite{goodfellow2014generative} framework describes a competitive game between two neural networks in which the generator network must compete against an adversary according to a game theoretic scenario \cite{goodfellow2016deep}. The generator produces samples from a noise distribution, and its adversary, the discriminator, tries to distinguish real samples from the generated samples. The real samples are inherited from the training data and the generated samples from the generator. % respectively samples inherited from the training data and samples produced by the generator. %By applying some of the Optimal Transport (OT) concepts gathered in \cite{villani2003topics} and noticeably, the Wasserstein distance, Arjovsky et al. introduced the Wasserstein GAN (WGAN) in \cite{arjovsky2017wasserstein} to improve the GAN's training. 
The Wasserstein GAN (WGAN) \cite{arjovsky2017wasserstein} facilitates the convergence of GANs' training using the optimal transport metric. The WGAN is further optimized by using a Gradient Penalty (GP-WGAN) method allowing the generation of adversarial samples of higher quality \cite{gulrajani2017improved}. With Wasserstein distance and the gradient penalty, the generator and the discriminator are able to improve at the same pace, avoiding mode-collapse a characteristic leading to unoptimized neural network weights which results in poor training. The GP-WGAN objective loss function is expressed such that

\begin{equation} \label{eq::DWGAN} 
L = \underset{\widetilde{X}\sim P_G}{\mathbb{E}}[f(\widetilde{X})] - \underset{X\sim P_X}{\mathbb{E}}[f(X)] + \lambda \underset{\widehat{X}\sim P_{\widehat{X}}}{\mathbb{E}} [(||\nabla_{\widehat{X}} f(\widehat{X})||_2 - 1)^2] 
\end{equation} 

where $f$ is the set of 1-Lipschitz functions on $(\mathcal{X}, d)$, $P_X$ the original data distribution, $P_G$ the generative model distribution implicitly defined by $\widetilde{X}=G(Z), Z\sim p(Z)$. The input $Z$ to the generator is sampled from a noise distribution such as a uniform distribution. $P_{\widehat{X}}$ defines the uniform sampling along straight lines between pairs of points sampled from the data distribution $P_X$ and the generative distribution $P_G$. A penalty on the gradient norm is enforced for random samples $\widehat{X}\sim P_{\widehat{X}}$.

\subsection{Distributed Denial of Service Attacks}
A DDoS attack occurs when multiple machines flood a network host with traffic until the host cannot respond, or crashes, preventing access to users \cite{douligeris2004ddos}. These attacks exploit protocols at the network, transport and/or applications layers. The well known attacks in the DDoS taxonomy include Teardrop, Smurf, UDP Flooding, SYN Flooding, NTP Flooding, DNS Amplification, and application layer HTTP attacks such as GoldenEye and Slowloris \cite{douligeris2004ddos}. In this work, we concentrate on generating synthetic network traffic for Smurf and GoldenEye attacks. In a Smurf attack, an attacker relies on a large collection of Internet Control Message Protocol (ICMP) echo request packets using a victim's spoofed source IP address that are broadcast in a network. In a GoldenEye attack, an attacker targets web servers by keeping connections %with them 
open %for a long as possible, 
with HTTP requests rendering the server unable to process any other requests. The GoldenEye attack and other similar application layer DDoS are deemed to be particularly intractable because of their inherent distributed nature.    

\subsection{SynGAN Framework and DDoS} 
\begin{figure}[b!]
 \centering
  \frame{\includegraphics[height=120pt,width=150pt]{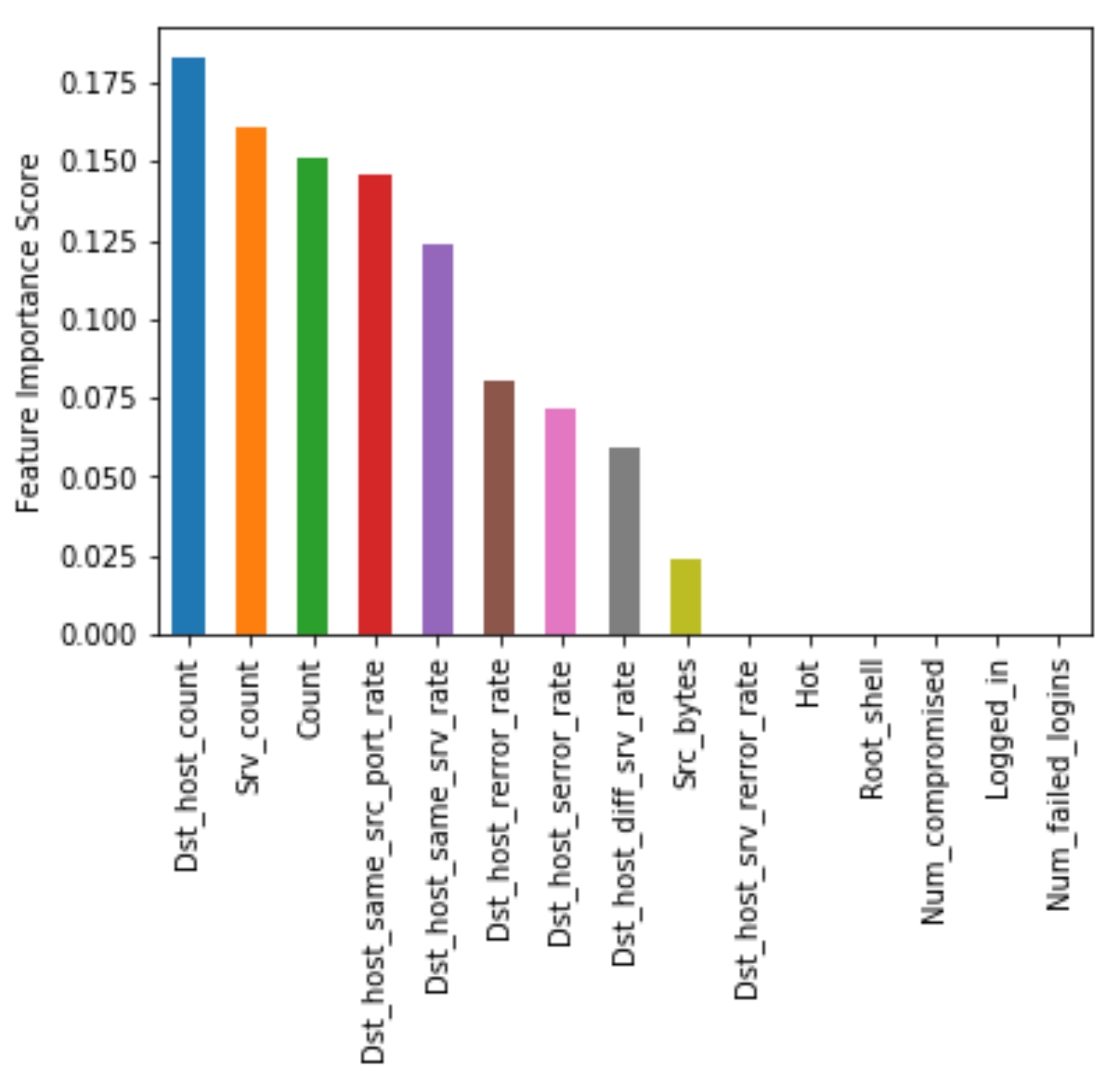}}
  \frame{\includegraphics[height=120pt,width=160pt]{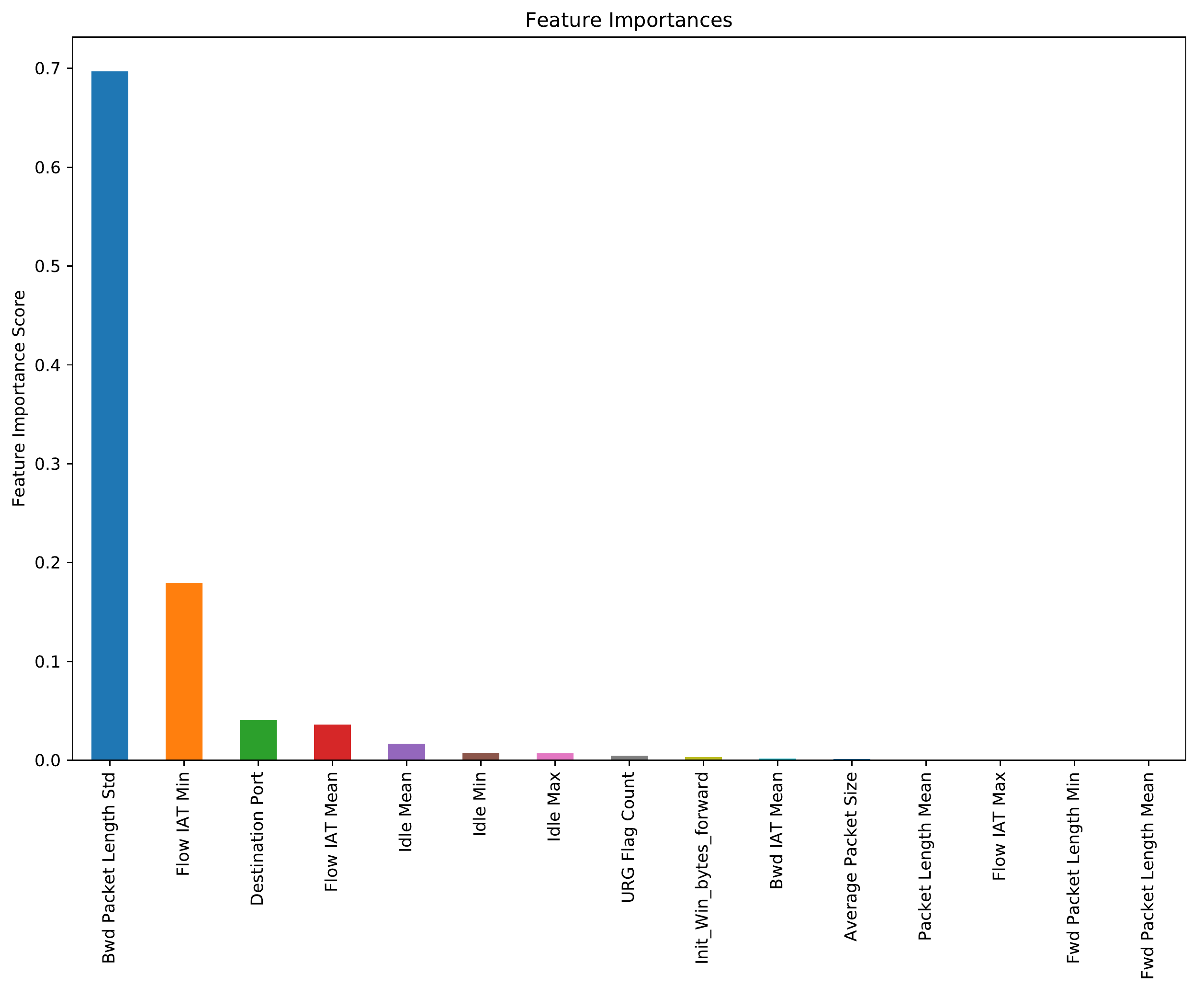}}
 \caption{From left to right. (a) Gradient boosting evaluator relevant features to detect NSL-KDD smurf attacks and (b) to detect CICIDS2017 GoldenEye attacks.}
 \label{fig1}
\end{figure}

The SynGAN framework generates synthetic network attack packets using real-world attack traffic mutations. It relies on the GP-WGAN formulation to ensure a better convergence of the error minimization function. As illustrated in figure \ref{fig::methodology}, the SynGAN framework has three components: the Generator, the Discriminator and the Evaluator. Initially, during the GAN's training, a uniform random distribution is used to initialize the artificial samples. Then, the generative network mutates the artificial samples, trying to make the latter similar to real attacks. Subsequently, the synthetic attacks are sent to the discriminator. The discriminator tries to differentiate between real and artificial attacks. It provides feedback to the generative network to recursively improve the quality of the generated adversarial samples. At the end of the GAN's training, only the generator is used to generate artificial DDoS attacks. Finally, the gradient boosting classifier, the evaluator, attempts to differentiate between between real and generated attack packets using the quality benchmark based on the root mean square error. We chose to use the gradient boosting algorithm as it allows an easy identification of the relevant features used for the traffic classification, as shown in figure \ref{fig1}. \\ %We describe the generation of artificial traffic with SynGAN in the algorithm \ref{algo1}. 

For our first attack generation implementation and analysis, we compare and evaluate the NSL-KDD %\footnote{The dataset is available at https://www.unb.ca/cic/datasets/nsl.html.} 
and CICIDS2017 %\footnote{The dataset is available at https://www.unb.ca/cic/datasets/ids-2017.html.} 
datasets. The NSL-KDD dataset contains two classes of network attacks, DoS and Probe, and two classes of host-based attacks, Remote-2-Local (R2L) and User-2-Root (U2R). We first start our approach on the DDoS attacks subset including the simple Smurf attack with 41 network flow features\cite{tavallaee2009detailed}. The CICIDS2017 dataset contains various attacks such as infiltration, heart-bleed and GoldenEye with 80 network flow features \cite{sharafaldin2018toward}. We then concentrate our synthetic attack experiments on the more complex GoldenEye attack pattern.

% !!! NEW SECTION !!! %
% !!! =========== !!! %
\section{Experiments} \label{sec::experiments}

The Keras \cite{chollet2015keras} library is used for the neural network implementation. The GP-WGAN has 5 layers with 256, 128, 128, 128 and 78 neurons, and a ReLU activation function for each of the layers. Additionally, we used the RMSProp gradient descent for the neural network training \cite{hinton2012rmsprop} with the parameters $\text{lr}=0.001, \rho=0.9, \epsilon=10^{-6}$. We empirically obtained the lowest error reconstruction with the gradient penalty $\lambda=10$. The simulations were performed on a computer with 16GB of RAM, Intel i7 CPU and a Tesla K80 GPU accelerator \cite{charlier2019}. %\footnote{The code is available at https://github.com/dagrate/ddosgan}. 
\\ 

\begin{figure}[b!]
 \centering
  \frame{\includegraphics[width=160pt]{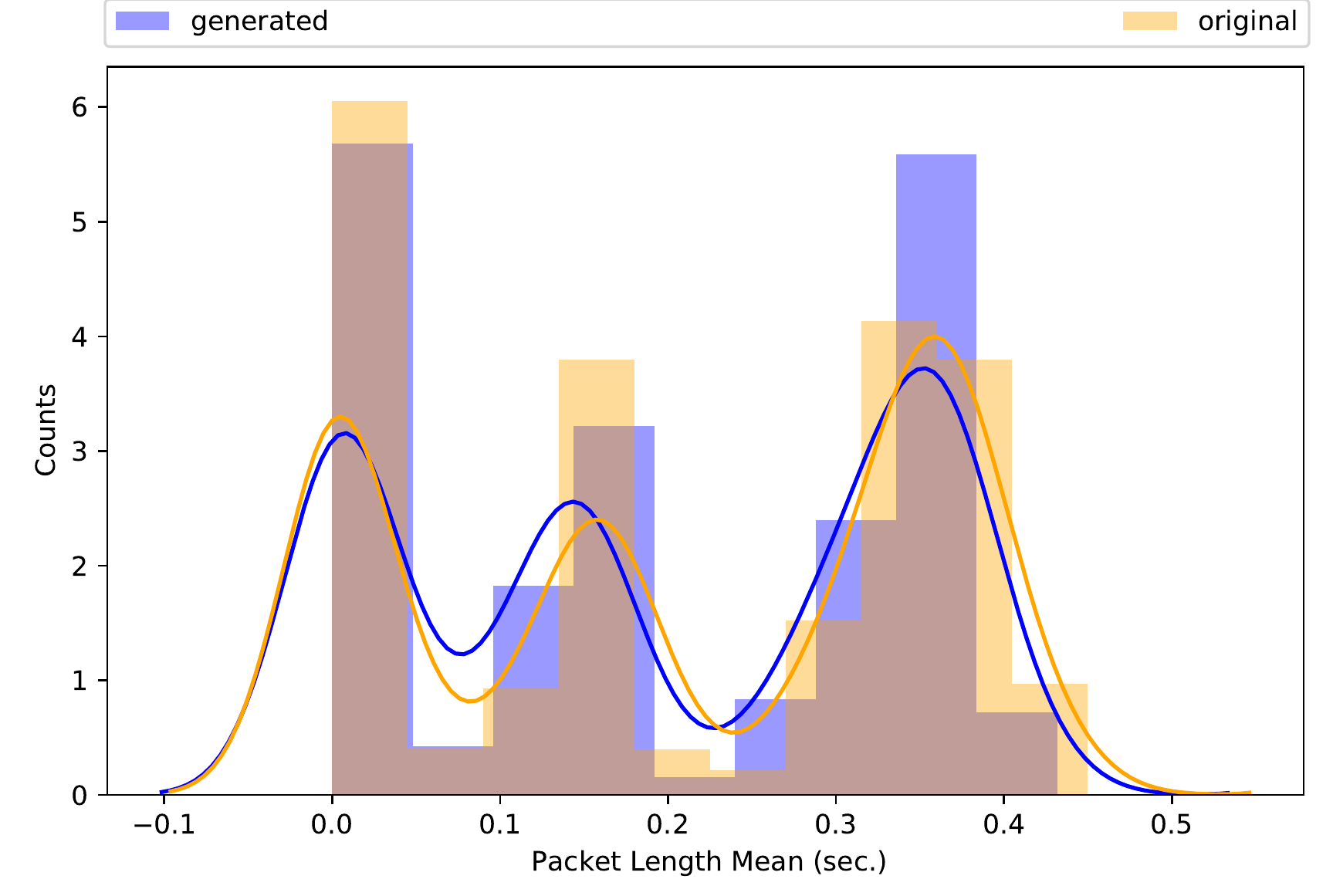}}
  \frame{\includegraphics[width=160pt]{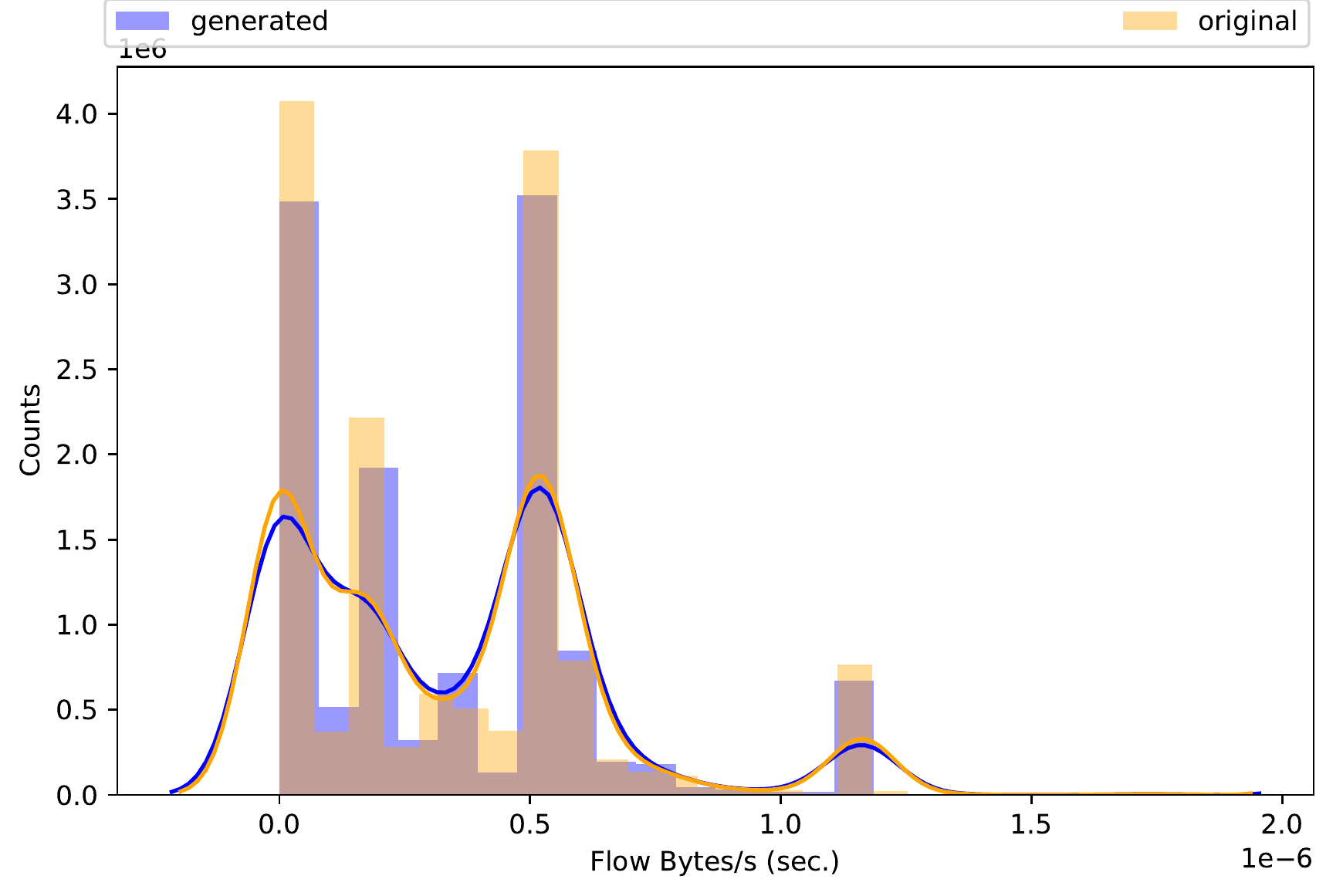}} \\ \vspace{.25cm}
  \frame{\includegraphics[width=160pt]{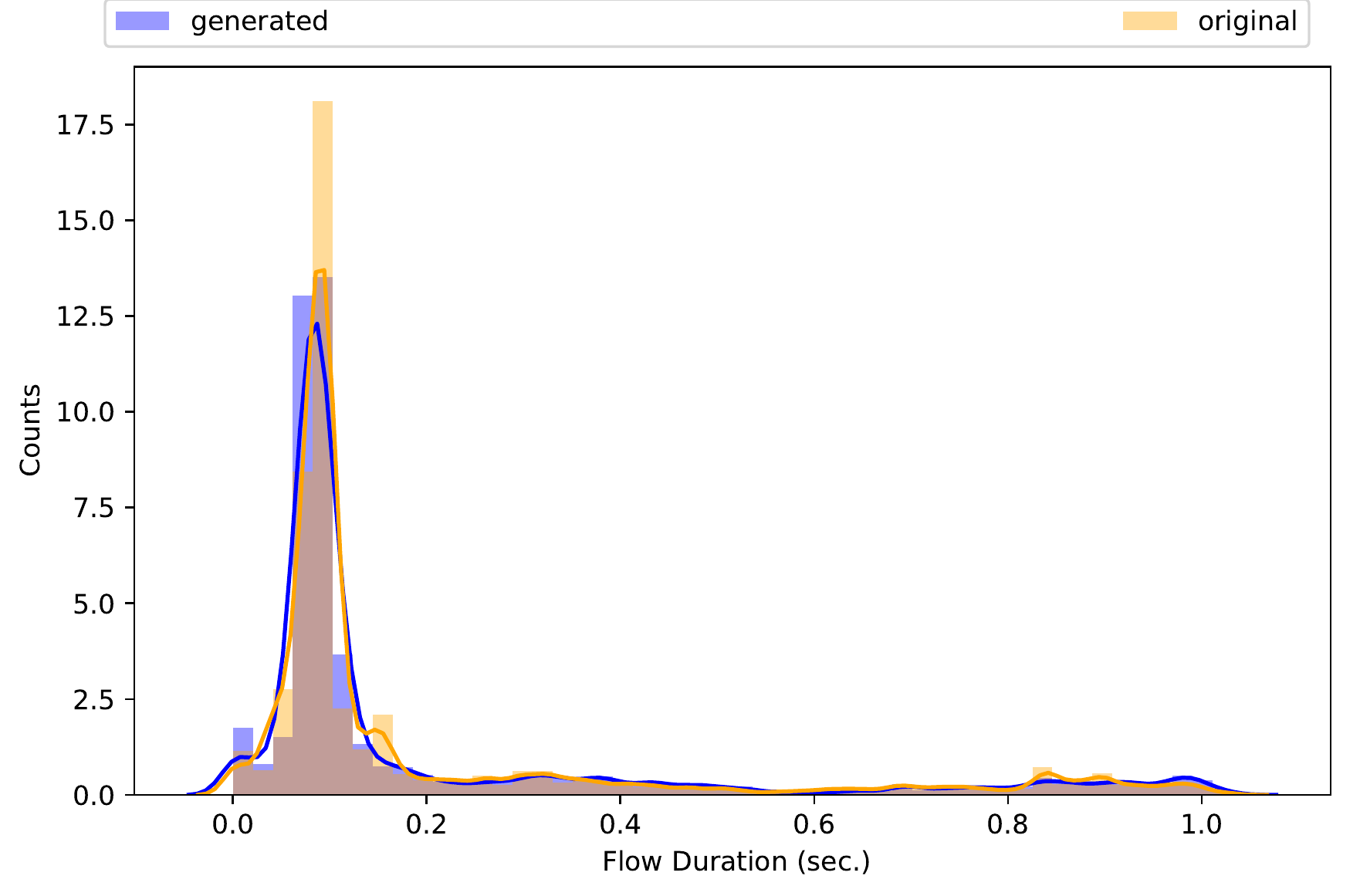}}
  \frame{\includegraphics[width=160pt]{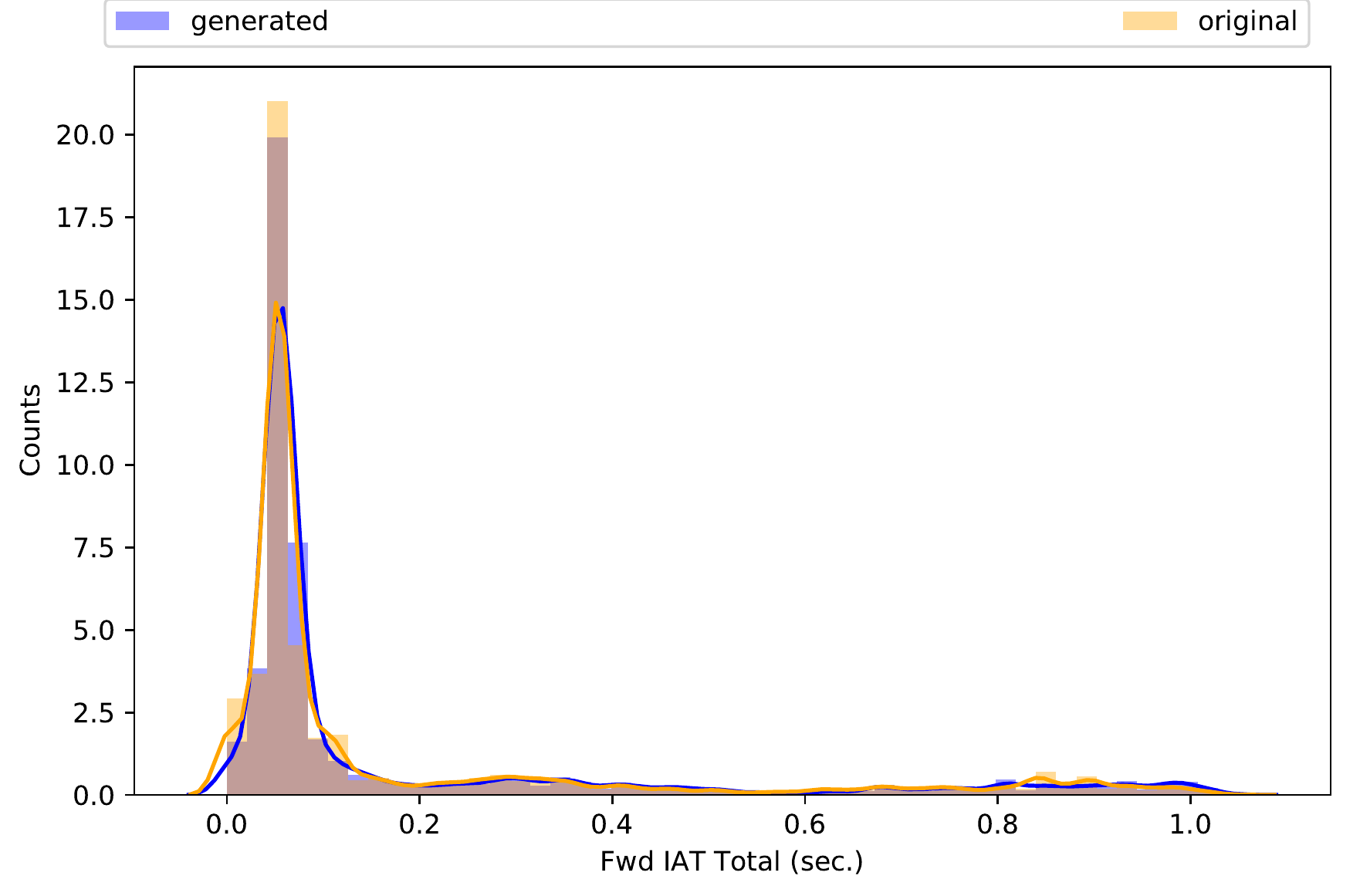}}
 \caption{From left to right. (a) Packet Length Mean, (b) Flow Bytes per sec.
 (c) Flow duration and (d) Forward Inter Arrival Time (IAT) distributions between the generated and the original attacks using SynGAN.}
 \label{fig2}
\end{figure}

\textbf{Results and Discussions} From the Smurf attack analysis of the NSL-KDD dataset, the top-15 aggregate parameters identified by the gradient boost evaluator are shown in figure \ref{fig1}. The highest weighted parameter, Dst\_host\_count, represents the number of connections having the same destination host IP address \cite{tavallaee2009detailed}. %The highest weighted parameter, Dst\_host\_srv\_count represents the number of connections having the same port number \cite{tavallaee2009detailed}. 
The NSL-KDD dataset, however, contains only 2,000 Smurf attacks resulting in weak statistical relevance, thus rendering our packet generation unreliable. We concentrated, therefore, on the CICIDS2017 dataset that has in excess of 10,000 DDoS attack samples including the GoldenEye attack. The SynGAN framework is able to generate adversarial attacks with a root mean square error of 0.10 implying very close similarity between the artificially generated attacks and original attacks. In figure \ref{fig2}, the graphs describing the packet length mean, the flow bytes per sec., the flow duration and the forward Inter Arrival Time (IAT) distributions between the generated and the original attacks confirm close similarity. The other features share the same distribution pattern. We computed the Area under the Curve (AUC) of the Receiver Operating Characteristic (ROC) curve of the gradient boosting classifier, the evaluator, between the generated and the true attacks. The preliminary AUC score is at 75\% in our experiments, highlighting that the evaluator is partly incapable to differentiate between the two types of attacks. %Future research will concentrate on achieving the target of 50\%, for which the evaluator will fully be incapable to differentiate between generated and true attacks. 
These results reaffirm the ability of the SynGAN framework to generate adversarial attacks of high quality.

%\begin{table}[t]
%\centering 
%\caption{TBU} \label{tab1}
%\scalebox{1.0}{ 
%\begin{tabular}{ccc} 
%  \toprule 
%  TBU & TBU & TBU  \\
%  \midrule
%  0 & 0 & 0  \\
%  0 & 0 & 0  \\
%  \bottomrule 
%\end{tabular}
%} 
%\end{table} 

% !!! NEW SECTION !!! %
% !!! =========== !!! %
\section{Conclusion} \label{sec::conclusion}

We introduced the SynGAN framework to generate synthetic network attacks using real attack traffic data. We evaluated the NSL-KDD and CICIDS2017 datasets for a subset of DDoS network attacks. The recent CICIDS2017 dataset has statistically more significant samples for DDoS attacks. 
%such as GoldenEye and Slowloris. 
The framework was applied to generate GoldenEye attack data showing more than adequate convergence of real and synthetic data. Although the current framework only generates DDoS network attacks using GP-WGANs, it shows promising results. Our future work will explore generating models having more complex state machines \cite{yu2017seqgan} to orchestrate different network attack types. The final goal is to evaluate the effectiveness of commercial NIDS by generating high-quality synthetic attack traffic  of different types, and integrate the framework in a network security automation pipeline. %We will also test the synthetic attacks against commercial IDS platforms.  
\bibliographystyle{./splncs04}
\bibliography{./zzz-mybibliography}
\end{document}